# Suggesting Relevant Questions for a Query Using Statistical Natural Language Processing Technique

Shriniwas Nayak[a], Anuj Kanetkar[a], Hrushabh Hirudkar[a], Archana Ghotkar[a] , Sheetal Sonawane[a] and Onkar Litake[a]

[a] *Computer Engineering, SCTR's Pune Institute of Computer Technology, Pune, India*

**Abstract: Background:** Suggesting similar questions for a user query has many applications ranging from reducing search time of users on e-commerce websites, training of employees in companies to holistic learning for students. The use of Natural Language Processing techniques for suggesting similar questions is prevalent over the existing architecture. Mainly two approaches are studied for finding text similarity namely syntactic and semantic, however each has its draw-backs and fail to provide the desired outcome. In this article, a self-learning combined approach is proposed for determining textual similarity that introduces a robust weighted syntactic and semantic similarity index for determining similar questions from a predetermined database, this approach learns the optimal combination of the mentioned approaches for a database under consideration. Comprehensive analysis has been carried out to justify the efficiency and efficacy of the proposed approach over the existing literature.

- **Objective:** This study tries to improve the selection and rejection ability for query recommendation system using a novel statistical method
- **Method:** This paper proposes a novel method to suggest similar queries for a user input query, by optimally combining syntactic and semantic similarity techniques, by calculating weighted arithmetic mean. The weight is data dependent and is calculated by the system.
- **Results:** Experimental results on 50K+ queries suggest that proposed novel method of combining syntactic and semantic systems is able to select similar queries as well as reject dissimilar ones, performing better than standalone syntactic or semantic systems.
- **Conclusion:** The comparison of syntactic and semantic approaches presented in the article helps to understand that both the methods suffer from drawbacks as discussed in the article, making the independent use of a single approach unsatisfactory. It can be learnt that a technique that combines both syntactic and semantic approaches performs better than the any standalone system. This Statistical method helps the user to interact with any machine in natural language with greater efficacy and ease.

## 1. INTRODUCTION

In today's world where Artificial Intelligence is ubiquitous, NLP (Natural Language Processing) as a sub domain has been able to prove its importance and wide range of applicability. NLP helps humans to interact with machines and receive responses from the machine that create an illusion of interacting with a human. Almost every field today involves some form of interaction between humans and machine, if this interaction is carried out in natural language or day to day language then it leads to greater comfort, the usability of the system increases and leads to a better and efficient interaction. Following is the structure of the rest of the paper. Section 2 surveys the related work. Section 3 indicates the different methods followed. Section 4 present the experimental setup. Section 5 explains our results. Section 6 lists the applications. Section 7 presents the conclusion of the paper.

### 1.1. Motivation

Lot of tasks in NLP are based on similarity [1] of sentences. In applications like document classification [2], plagiarism [3], summarization [4], sentiment analysis [5], question answering [6] similarity plays an important role. Mainly two similarity approaches have been discussed over the existing literature, namely syntactic [7] and semantic [8]. The syntactic approach computes similarity on word level similarity, whereas semantic approach computes similarity based on meaning of words present in text. These approaches have certain draw-backs as discussed in the literature further which suggest that either cannot serve the purpose of suggesting similar questions when used independently.

In today's technologically advancing world, getting answers for questions, general or specific is very important to the development of the user and as a result, development of the overall community eventually. Thus, this article proposes to find the relevant questions to a query entered by the user using a combined syntactic and semantic approach that determines the optimal weight to be as-signed to an approach using a statistical rank-based technique. to solve doubts thereby enhancing knowledge of the user as compared to a standalone approach.

## 2. RELATED WORK

Textual Similarity approaches is broadly classified into two types Syntactic and Semantic Similarity. Syntactic Similarity deals with just the words of the sentences without going into its meaning for example Cosine Similarity and n-grams, whereas Semantic similarity takes the meaning of the words into consideration for example Knowledge Based Similarity [9]. Syntactical Similarity is understood by studying its methods such as n-grams syntactical similarity which computes the similarity between two sentences using Tree Edit Distance and Cosine similarity which computes the similarity using the cosine formula [10]. Semantic Similarity is studied in two parts. But first we need to understand what Semantic similarity is and why it is required. We then look at the basic Semantic Similarity approaches which are Corpus Based and Knowledge Based Similarity. Corpus Based and Knowledge Based Similarity are used for Semantic Similarity by using techniques such as Point wise Mutual Information and Latent Semantic Analysis [11] [12]. We then look at a Semantic similarity approach which uses a concept from Syntactic similarity. The method is Sparse Vector Densification method which uses Cosine Similarity by making some changes to its formula and changing the vectors in which data is stored [13]. We then look at a new and better way for Semantic Similarity which is a combination approach. The approach uses LSA with WordNet knowledge set to per-form Semantic Similarity. It compares the words according to the eight defined rules which helps in similarity calculation using LSA. This provides a better and efficient way for Semantic Similarity which is shown by its results in English and Spanish language [14]. Till now we have discussed Syntactic and Semantic similarity separately. But there is also a combined approach for syntactic and semantic similarity that uses a tree kernel based approach for combining the two approaches [15]. A method that calculates a weighted mean of similarity scores calculated by using both the approaches has also been introduced. However, the mentioned articles present the results obtained with different combinations but no comment on selecting the best combination is made. The choice of the same is to be done manually, this process is tedious and lengthy. This article aims at bridging this gap by suggesting a method that self learns the optimal combination of syntactic and semantic methods [16].

## 3. METHODS

This sections provides a detailed discussion on syntactic and semantic approaches and compares them. This section also introduces the methodology proposed for learning the optimal combination of syntactic and semantic approaches.

### 3.1. Syntactic Similarity Approach

Syntactic similarity approach is based on the word level similarity. Given a query or block of text to be compared with another, this approach finds the number of common words occurring in both and assign a similarity score based on the same. Cosine similarity approach has been used in this article. Given a query or block of text it is first preprocessed. Preprocessing includes removing punctuation, converting all words to lower case, removing stop words i.e. removing words which carry very less or no significance, example of such words are "a", "the", "is" amongst others. After this basic preprocessing tokenization and stemming [17] is performed. This helps to represent the text as a vector. Both the given text from database and the user input text are represented as vectors. The cosine similarity formula is used to determine the similarity amongst these two vectors.

$$\cos(t,e) = \frac{te}{\|t\|\|e\|} = \frac{\sum_{i=1}^{n} t_i e_i}{\sqrt{\sum_{i=1}^{n}(t_i)^2}\sqrt{\sum_{i=1}^{n}(e_i)^2}} \quad (1)$$

The similarity score ranges from 0 which represents no similarity and such vectors are called as orthogonal vectors to 1 which represents complete similarity and such vectors are known as perfectly aligned vectors.

### 3.1.1. Time Complexity Analysis of Syntactic Approach:

The time complexity analysis of syntactic approach includes the time required for preprocessing and the time required for computing the similarity using equation 1. The time complexity is calculated considering that there are n queries in the database and similarity is to be computed with respect to a single user input query. Time complexity required for each step namely punctuation removal, lower case conversion, tokenization and stop words removal is $O(n)$, where n represents the number of queries in the database. The time complexity required for the process of stemming is $O(K)$ [18]. Similarity score computation using cosine formula is done in constant time and time complexity can be represented as $O(K1)$ This process id repeated for n queries in the data set. The overall time complexity for the syntactic similarity approach is represented in equation 2.

$$TC(S_1) = n * (4 * n + K + K1) = O(n^2) \quad (2)$$

Where S1 stands for Syntactic System, n represents the number of queries in the database

### 3.2. Semantic Similarity Approach

Semantic similarity tries to find the similarity score based on the meaning of the words in the query or text. The model used for this purpose is Doc2Vec [19], the underlying model is Word2Vec [20]. Word2Vec is implemented using two main methods namely continuous bag of words and the the skip gram model [21], in this article the continuous bag of words model is used. The Doc2Vec model is an extension of the Word2Vec model wherein a vector contains numeric representation of the document. Distributed Memory version of Paragraph Vector (PV-DM) has been considered. This method tries to find out words which are missing from the context, it adds these words and then uses cosine similarity method to generate a similarity score. This method involves creating set of all words that appear in the document which is referred as list of unique words. This list helps the word2vec model to learn associations efficiently.

### 3.2.1. Time Complexity Analysis of Semantic Approach:

The time complexity analysis of the semantic approach includes the time required for preprocessing and finding the semantic similarity score. The time complexity required for each step namely punctuation removal, lower case conversion, tokenization, and stop words removal is O(n), where n represents the number of queries in the database. The time required for calculating the similarity score can be represented as O(g(V))(V)) [22], where n represents the number of queries in the database and V represents the size of the word list created containing unique words. This process is repeated for every query in the database. The overall time complexity of the semantic approach is represented in equation 3.

$$TC(S_2) = n * (n * \log(V) + 4 * n) = O(n^2 * \log(V)) \quad (3)$$

Where $S_2$ stands for Semantic System, n represents the number of queries in the database and V represents the size of unique word list created.

### 3.3. Comparison of Syntactic and Semantic Approach:

Both the approaches discussed earlier suffer from certain drawbacks as discussed further. Both the approaches were implemented on a database with about thirty two thousand queries. Analysis regarding the distribution of frequency of queries with respect to the similarity percentage is presented. Figure 1 represents the distribution by using semantic approach and figure 2 represents the same using semantic approach. It can be learnt from the figures that syntactic approach represents majority of the vectors as near orthogonal vectors whereas semantic approach does not. It can be inferred that it is easier to decide a similarity threshold for syntactic approach as compared to semantic approach.

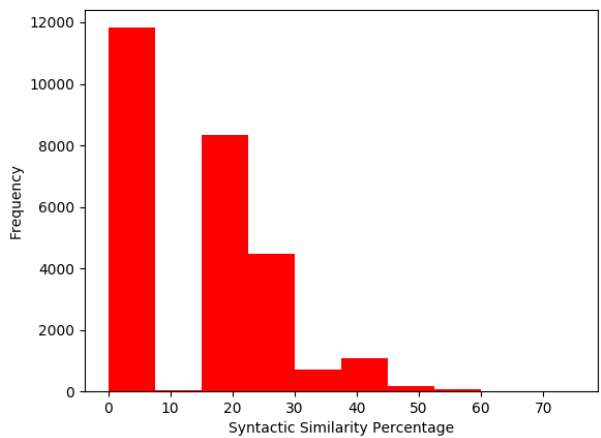

**Fig. (1).** Similarity Score Bar Graph using Syntactic Approach

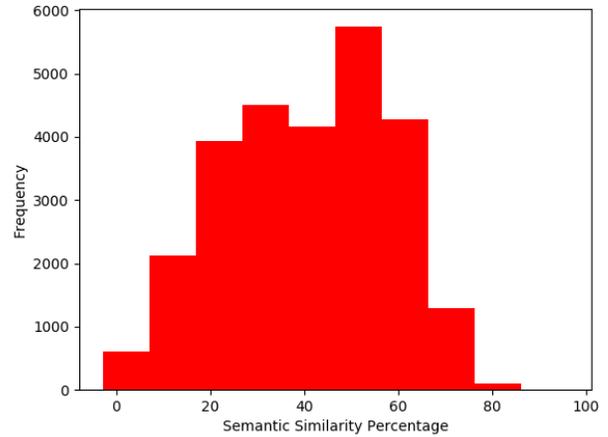

**Fig. (2).** Similarity Score Bar Graph using Semantic Approach

The summary of the comparative analysis of both the similarity approaches can be studied with respect to these points Performance: Syntactic system requires less computational power whereas semantic system may require a Graphics Processing Unit for interactive usage Drawbacks: Syntactic system represents a considerable majority of queries as orthogonal vectors which may result into deletion of semantically similar queries, for example" What is your age?" and" How old are you?" are represented as orthogonal vectors even though they mean the same. On the other hand, Semantic system has poor rejection ability, consider two sentences "How to read string in java" and "Syntax to read text in python", in an application focusing on helping new learners to study only one programming these two sentences should be considered dissimilar. A semantic system would categorize them as almost co-linear vectors due to similar meaning, indicating that at times semantic system lacks ability to reject dissimilar sentences. Syntactic system has poor acceptance rate and Semantic system has poor rejection rate, proposed algorithm overcomes these drawbacks. Syntactic system is better suited for key word intensive queries and Semantic system will provide better performance on queries involving natural language.

### 3.4. Self Learning Combination Approach:

To overcome the drawbacks of both the approaches mentioned above and to use the benefits of both a combined approach can be used. Using both the approaches similarity score is calculated and to obtain the final similarity score weighted arithmetic mean of these scores is calculated. The proposed approach learns this optimal weights to be assigned to each of the methods using a rank based approach. This eliminates the need of manually selecting the best weight and helps to efficiently combine both the approaches. The combination of the approaches has been represented in equation 4.

$$\text{Combined\_simalarity} = \lambda * \text{S1} + (1 - \lambda) * \text{S2} \quad (4)$$

Where S1 represents similarity score calculated using Syntactic approach and S2 represents similarity score computed using Semantic approach.

The value of λ ranges from 0.0 to 1.0, as can be learnt from equation 4 when the value of λ is 0.0 the system behaves as completely semantic and when λ is equal to 1.0 the system is behaves as a standalone syntactic system. The optimal value of λ will differ for each database, it depends on the type of queries present in the database. If the queries are key word intensive then the value of λ will tend to shift towards 1.0 and if the queries present the use of natural language then λ will be close to 0.0.

The proposed approach receives list of queries from the database with ranks provided to them by an admin with respect to one user input query. Ranks near to one indicate that the query is similar and system should accept it whereas ranks with higher value indicate that these queries are dissimilar and system should reject them. After receiving such a database i.e. some or all queries ranked according to a user input query, the system calculates both syntactic and semantic similarity scores. For different values of λ ranging from 0.0 to 1.0 in redefined steps, the final similarity score is calculated as per equation 4. The queries are sorted on the basis of this combined score, the query with maximum score is given a rank one and so forth. The square of difference of the rank assigned by the admin and the rank generated by the user for a particular value of λ is calculated, this value is calculated for all the queries assigned rank by admin. This value is called as SSRD (Sum of Squared Rank Difference). Multiple values generated using multiple values of λ are compared and the value of λ for which SSRD is minimum is selected as the optimum or best value. Figure 3 provides a representation of the self learning technique.

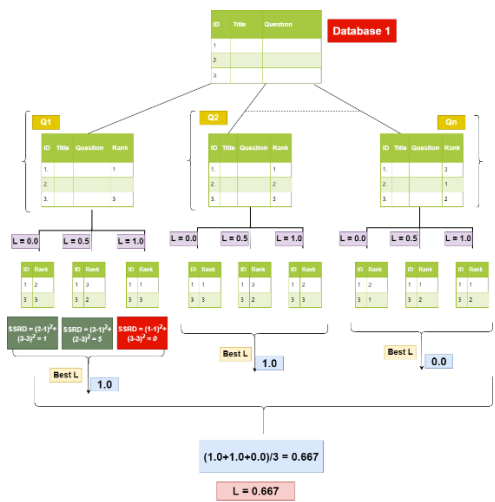

**Fig. (3).** Similarity Score Bar Graph using Syntactic Approach

The database 1 tagged in figure 3 can be considered as the parent database, using this database multiple files are created and ranks are assigned to some or all queries in the database. This file is referred as the ranked file, one user query and corresponding ranked file form a single set. Such multiple sets are mentioned in driver file. The detailed process for computing the optimal value of λ that is to determine the optimal weights to be assigned for syntactic and semantic approaches is explained in algorithm 1

**Algorithm 1:** Find best weights for syntactic and semantic approach

FIND_OPTIMAL_λ(DRIVER_FILE)
1:   initialize step_size := 0.1
2:   M := number of (query, ranked_file) sets in driver_file
3:   **while** driver_file is not traversed **do**
4:       read query, ranked_file from driver_file
5:       S1 := Syntactic_Similarity(query, ranked_file)
6:       S2 := Semantic_Similarity(query, ranked_file)
7:       initialize λ := 0.0
8:       **while** λ ≤ 1.0 do
9:           **while** ranked_question is present in ranked_file **do**
10:              combined_similarity := λ ∗ S1 + (1 − λ) ∗ S2
11:          sort list_of_questions, key : combined_similarity
12:          assign rank, order : decreasing
13:          initialize SSRD := 0
14:          **while** ranked_question is present in ranked_file **do**
15:          SSRD:= SSRD + (predicted_rank-assigned_rank)$^2$
16:          store(λ, SSRD)
17:          λ := λ+ step_size
18:      bestλ_for_set := λ where SSRD is minimum
19:      store(bestλ_for_set)

$$\frac{\sum_{i=1}^{m} best\lambda\_for\_set[i]}{m}$$

20:   optimalλ :=
21:   **return** optimalλ

The driver file as mentioned in algorithm 1 contains multiple sets, wherein one set constitutes a user query and a name of file containing some or all questions ranked according to the query. This driver file is provided as input to algorithm 1, which learns best weights for one particular set. Average of these values is computed in order to obtain the final optimal weight referred as optimalλ. This value is applicable for the complete database. The learnt value of λ provides better results if learning is conducted using multiple sets, however the system can learn the optimal using a single set as well.

*3.4.1. Time Complexity for determining Optimal weights:*

Time complexity analysis of learning the optimal wights can be divided into three parts, namely syntactic similarity computation, semantic similarity computation and calculation the value of ssrd (sum of squared rank difference) for multiple values of λ. Time complexity for syntactic similarity can be obtained from equation 2 and for semantic similarity can be obtained from equation 3. The calculation of best lambda occurs in constant time and this process is repeated m times, where m represents the number of (query, Ranked File) sets provided in the driver file. The over all time complexity for learning the optimal weights is represented in equation 5.

$$TC(Learning\_Optimal\lambda) = m * (TC(S_1) + TC(S_2)) = O(m * n^2 * \log(V)) \quad (5)$$

In equation 5 S1 and S2 denote syntactic and semantic similarity as represented in equation 2 and 3, m represents the number of (query, Ranked File) sets in driver file, n represents the number of questions in the database and V holds the size of unique list of words created. The algorithm learns the optimal value with time complexity as per equation 5, this time complexity is required only while training the model, the time complexity of determining the similarity for a user query, once the model has been trained is represented in equation 6, which is same as time complexity required for a semantic system. S$\lambda$ represents the system using the proposed optimal lambda approach to determine similarity.

$$TC(S\lambda) = (TC(S1) + TC(S2)) = O(n^2 * \log V) \quad (6)$$

### *3.4.2. Analysis of Lambda Parameter:*

The value of Lambda Parameter ($\lambda$) varies from 0.0 to 1.0 and is data dependent. Semantic similarity approach performs better on data sets containing wide usage of synonyms or similar phrases; for example, e-commerce portals. Syntactic similarity approach is better for data sets containing limited vocabulary, for example, teaching portals. Hence, the value of ($\lambda$) tends to be closer to 0.0 as per equation 5, indicating dominance of syntactic similarity score, in case of data sets with limited vocabulary; whereas value closer to 1.0 suggests that data set consists of rich vocabulary and semantic score is assigned higher weight.

### **3.5. Self Learning Combination Approach:**

Data flow and processing methodology from a system architectural point of view are represented in figure 4. The end user interacts with the system by submitting an input query; this input query acts as the input for the processing back end, which using pre-computed value of lambda (value indicating learned weights for syntactic and semantic system) computes similarity for queries present in the database.

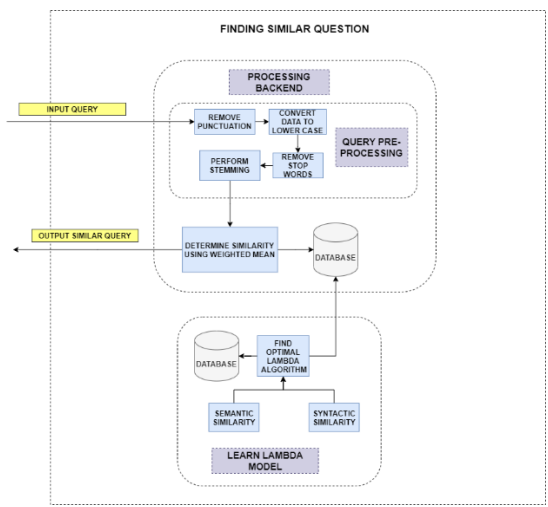

**Fig. (4).** System Flow Architecture Diagram

Queries from the database are ranked as per the similarity score, which is computed using the optimal value of lambda. Queries with higher similarity score are suggested to the decided by the value of a pre-determined threshold on the similarity score or on the number of queries to be displayed. The user input query is first pre-processed using known techniques like punctuation removal, changing all text data to lower case, removing stop words and stemming the words. This pre-processed query is vectorized and then fed to the similarity calculation process. This process reads the optimal value of lambda stored in the database. The learn lambda model learns the optimal value of lambda based on ranking provided by admin and stores intermediate as well as the final optimal value of lambda in the database. The model uses both syntactic and semantic similarity techniques to compute the optimal value of lambda as represented in Algorithm 1.

### **4. EXPERIMENTAL SETUP AND DATA SOURCES**

The analysis and experimentation mentioned in the article have been performed with Python 3.6.9. The implemented codes use Nltk (Natual Language Tool Kit) 3.4.5 [23] library for natural language processing tasks, Matplotlib 3.1.1 [24] library for creating graphs, and Pandas 0.25.1 [25] library for reading the data. All data sets used for analysis of the open-source have the a of the presented approach have been obtained through open-source resources, the principle data set is available on Kaggle [26]. For the implementation of Word2Vec Model binary file made available by google [27] has been used. Data sets considered include text queries from a variety of fields; that include but are not limited to a, technical domain, e-commerce domain and general search engine-based queries. The data sets in total consist of queries over 100,000.

The principle data set is available on Kaggle [27], consisting data points in excess of 25 thousand. Each data point represents question submitted on stackoverflow portal by anonymous users which contains title of the question, body of the question and timestamp representing date and time of posting.

### **5. RESULTS**

The proposed methodology to learn optimum weightage has been implemented with multiple data sets with varied characteristics. The value of ssrd (sum of squared ranked difference) is used to asses the performance of a particular value of $\lambda$ as suggested in equation 4. The value of ssrd should be as minimum as possible and ideally 0 because ssrd indicates the inability of the system to predict the rank provided by the admin. The plot of value of ssrd against the value of $\lambda$ helps to understand the effect of different combinations of syntactic and semantic approaches.

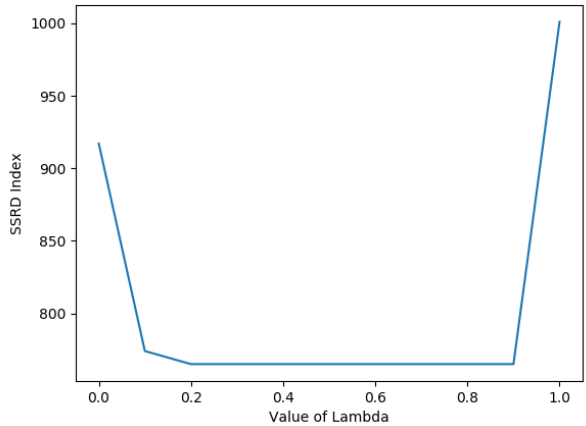

**Fig. (5).** Sum of Squared Rank Difference versus λ

It can be observed from figure 6 that the value of ssrd is high at extremities of λ that is 0.0 and 1.0. The reason for this is that when value of λ is near 0.0 the system behaves as purely semantic and is unable to reject dissimilar sentences. On the other hand when value of λ is near 1.0 the system behaves as a standalone syntactic system, lacking the ability to accept semantically similar sentences.

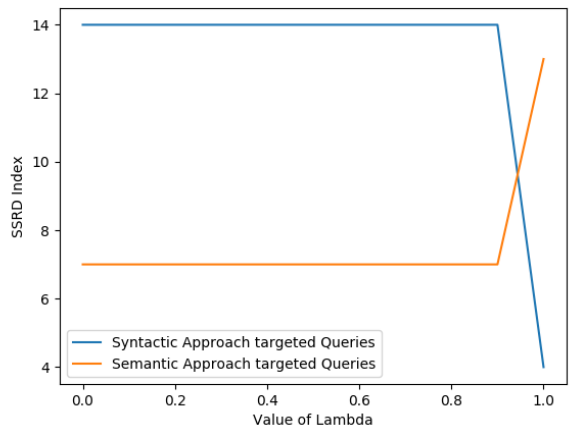

**Fig. (6).** Analysis of SSRD index

Figure 6 represents the analysis of ssrd index in two particular cases. Algorithm 1 was implemented on data sets, where data set one had keyword intensive queries referred as syntactic approach targeted and the second consisted of natural language queries referred as semantic approach targeted queries. It can be observed from figure 6 the value of ssrd jumps at the extremities for both the data sets, this implies that the proposed approach can also identify data sets for which standalone systems using either of syntactic or semantic approach produce expected results.

To assess the accuracy of three approaches namely syntactic, semantic and combined the ssrd score has been used. Firstly the ssrd score is calculated with system that suggests similar queries by making random choice, then ssrd scores are calculated for each method. The percentage reduction in ssrd score for a method indicates the effectiveness of that method, ideally the error reduction percentage should be 100.

| Data Set Type | S1 | S2 | S3 | Optimal λ |
|---|---|---|---|---|
| Mixed | 67.96 | 70.65 | 75.52 | 0.2 |
| Syntactic | 77.42 | 54.83 | 77.42 | 1.0 |
| Semantic | 55.17 | 75.86 | 75.86 | 0.0 |

In table I S1 indicates Syntactic Approach, S2 indicates Semantic Approach and Sλ represents combined approach. The value of λ has been computed using algorithm 1.

It can be inferred form table I that, error reduction ability and hence the efficacy of optimal λ method is better as compared to syntactic and semantic approaches for databases containing queries with keywords as well as natural language. If a particular database has only syntactic i.e. keyword intensive queries or only semantic i.e. natural language queries, then the system's performance is same as that of a standalone syntactic or semantic system respectively. This learnt value of λ is applicable to the complete database and may be required to be recomputed if there is significant alteration in the database. Domain expert or admin can run periodic tests for dynamic databases.

Plots represented in figure 4 and 6 indicate that ssrd (sum of squared rank difference) changes with the value of λ significantly enough in order to determine the optimal value of λ. The results indicate that the proposed method successfully determines the optimal weights to be assigned to syntactic and semantic approach.

The results indicate that the proposed method successfully determines the optimal weights to be assigned to syntactic and semantic approach.

## 6. APPLICATIONS

The methodology presented in this article can be used in any filed involving human and machine interaction. The proposed technique is generic and does not require any modification for implementation for a particular domain. Some prominent applications of this approach are discussed further. In healthcare domain predicting queries that may arise in near future depending on the present query helps to spread awareness and society at large will be better informed about dis-eases and epidemics, using optimal combination of syntactic and semantic approaches this can be achieved. Use of self-learning algorithm will ensure that only required queries are displayed to the user. For learning portals online courses and e-teaching methodologies are gaining importance at enormous rates. The proposed approach can be embedded with a chat bot that enables learners to ask questions as they would to an instructor. Use of the proposed methodology will help to explore new topics by suggesting related queries and lead to holistic development. In E-Commerce websites the presented approach will help to significantly reduce the search time for probable consumers on e-commerce websites. Consider a user requests for a particular commodity, the system under consideration will suggest similar products and help to resolve issues related to the commodity in natural language. For company training every company or firm hires new employees on a regular basis, a daunting task for the companies to train the freshers. The proposed system will help immensely by suggesting queries that the employee may encounter in near future. This helps the new employee to get accustomed to the new workplace efficiently

# 7. CONCLUSION

The comparison of syntactic and semantic approaches presented in the article helps to understand that both the methods suffer from drawbacks as discussed in the article, making the independent use of a single approach unsatisfactory. It can be learnt that a technique that combines both syntactic and semantic approaches performs better than the any standalone system. The results prove that the proposed statistical method based on ssrd (sum of squared rank difference) index can self learn the optimal weightage to be used for each approach. This Statistical method helps the user to interact with any machine in natural language with greater efficacy and ease.

Future scope of this methodology includes extending this approach of optimally combining rule-based approach, machine learning-based approach and deep learning-based approach optimally to improve the efficacy of other natural language processing tasks like text summarization, named entity recognition, sentiment analysis amongst others. The future scope also includes training the pro-posed model for languages other than English.

It is worth noting that the proposed algorithm is generic in nature and does not require any modification for implementation in diverse fields. Using a combined approach with optimum weights assigned to each approach helps to incorporate the benefits of both the approaches. The proposed system has is able to select similar questions as well as reject dissimilar questions. The proposed algorithm learns the best weightage for combined similarity score computation eliminating the tedious manual procedure required to do so thereby improving the efficacy and efficiency of a combined system.